\documentclass[10pt,twocolumn,letterpaper]{article}

\usepackage[pagenumbers]{cvpr} 

\usepackage{graphicx}
\usepackage{amsmath}
\usepackage{amssymb}
\usepackage{booktabs}
\usepackage{multirow}

\usepackage[pagebackref,breaklinks,colorlinks]{hyperref}

\usepackage[capitalize]{cleveref}
\crefname{section}{Sec.}{Secs.}
\Crefname{section}{Section}{Sections}
\Crefname{table}{Table}{Tables}
\crefname{table}{Tab.}{Tabs.}

\def\cvprPaperID{*****} 
\def\confName{CVPR}
\def\confYear{2023}

\begin{document}

\title{1st Place Solution for the 5th LSVOS Challenge: Video Instance Segmentation}

\author{Tao Zhang$^{1}${\qquad}Xingye Tian$^{2}${\qquad}Yikang Zhou$^{1}${\qquad}Yu Wu$^{1}${\qquad}Shunping Ji$^{1}$\thanks{ Corresponding author.}\\
Cilin Yan$^{3}${\quad}Xuebo Wang$^{2}${\quad}Xin Tao$^{2}${\quad}Yuan Zhang$^{2}${\quad}Pengfei Wan$^{2}$ \vspace{3mm}\\
$^{1}$Wuhan University\qquad$^{2}$Y-tech, Kuaishou Technology\qquad$^{3}$Beihang University
}
\maketitle

\begin{abstract}
Video instance segmentation is a challenging task that serves as the cornerstone of numerous downstream applications, including video editing and autonomous driving. In this report, we present further improvements to the SOTA VIS method, DVIS. First, we introduce a denoising training strategy for the trainable tracker, allowing it to achieve more stable and accurate object tracking in complex and long videos. Additionally, we explore the role of visual foundation models in video instance segmentation. By utilizing a frozen VIT-L model pre-trained by DINO v2, DVIS demonstrates remarkable performance improvements. With these enhancements, our method achieves 57.9 AP and 56.0 AP in the development and test phases, respectively, and ultimately ranked \textbf{1st} in the VIS track of the 5th LSVOS Challenge. The code will be available at \href{https://github.com/zhang-tao-whu/DVIS}{https://github.com/zhang-tao-whu/DVIS}.
\end{abstract}

\section{Introduction}
\label{sec:intro}

Video instance segmentation is a challenging task that extends the concept of image instance segmentation to videos. The objective of video instance segmentation is to simultaneously classify, track, and segment all instances of interest in a video \cite{masktrackrcnn}. It serves as a fundamental component for several downstream tasks, such as video comprehension, video editing, and autonomous driving.

Recently, there has been increasing attention on the performance of video instance segmentation methods in real-world scenarios. While classic offline methods such as Mask2Former-VIS \cite{mask2formervis}, SeqFormer \cite{seqformer}, IFC \cite{ifc}, and VITA \cite{vita} have been proven effective for short videos with simple scenes \cite{vis}, they often struggle to perform well on long videos with complex scenes \cite{ovis}. DVIS \cite{DVIS} thoroughly analyzed this problem and concluded that achieving end-to-end modeling of instance representations throughout the entire video is extremely challenging, which is the fundamental reason for the aforementioned phenomenon. To address this challenge, DVIS has devised a solution that decomposes video instance segmentation into three sub-tasks: segmentation, tracking, and refinement. Additionally, DVIS has introduced the referring tracker and temporal refiner to improve instance tracking stability and enhance information utilization, respectively.

The learnable referring tracker proposed by DVIS demonstrates exceptional tracking performance and significant potential when compared to heuristic association algorithms. However, there is still significant room for improvement in the referring tracker. Currently, the tracker utilizes the instances queries matched by heuristic algorithms as input, which is considered to include noise. However, in most scenarios, these inputs are already accurate and only contain negligible noise. Consequently, the tracker tends to converge to the shortcut. To fully unleash the potential of the learnable tracker, we believe it is crucial to enhance the task's difficulty by introducing noise to the input. This will enable the tracker to develop a stronger instance tracking ability.

In this report, we introduce a denoising training strategy for enhancing the performance of the referring tracker. Specifically, pronounced noise is intentionally incorporated into the instance query, serving as the tracker's input. During the course of training, the tracker is compelled to learn the strategic removal of this noise. We proffer three distinct noise simulation strategies, namely: weighted averaging, random cropping coupled with concatenation, and random shuffling. Notably, empirical evidence supports that the employment of each aforementioned noise strategy indeed advances performance. Remarkably, the random shuffling strategy demonstrates superior efficacy in bolstering performance, as it faithfully replicates the challenging scenarios typically encountered during the inference stage. Furthermore, the positive augmentative effect of the denoising training strategy on performance is further promoted by elongating the training iterations. Conversely, when denoising training strategy is not incorporated, models show no improvement in performance, even under extended training durations.

We also explore the effect of integrating the visual foundation model into video instance segmentation. Specifically, we employ the frozen ViT-L \cite{VIT} model, pre-trained with Dino V2 \cite{dinov2}, to offer more robust visual features for DVIS. By incorporating the visual foundation model, we enhance the discriminative ability of the instance representations generated by the segmenter. Consequently, the segmentation and tracking performance of DVIS experience a significant improvement.

The aforementioned improvements have resulted in the enhanced performance of DVIS, yielding noteworthy results on the YouTube 2021 dataset \cite{vis} and OVIS dataset \cite{ovis}, with AP scores of 64.6 and 53.9, respectively. Moreover, DVIS demonstrated its superior performance in the VIS track of the 5th LSVOS competition by ranking 1st place in both the development and test stages, with AP scores of 57.9 and 56.0, respectively.

\section{Method}

In this section, we will introduce the technical details. Since the overall network architecture is the same as DVIS, a detailed explanation will not be repeated here. Please refer to DVIS for detailed information regarding the model. The proposed denoising training strategy will be discussed in \cref{sec:noise}, while the utilization of visual foundation model will be introduced in \cref{sec:vfm}.

\subsection{Denoising Training Strategy} \label{sec:noise}
The objective of the proposed denoising training strategy is to intensify the challenge of the task by introducing simulated noise into the tracker’s input. This approach enables more efficient training of the tracker and enhances its capability for object tracking. At the heart of this strategy lies the noise simulation approach. In this paper, we present three noise simulation strategies, which include weighted averaging, random cropping coupled with concatenation, and random shuffling.

\begin{figure}[t]
  \centering
   \includegraphics[width=1\linewidth]{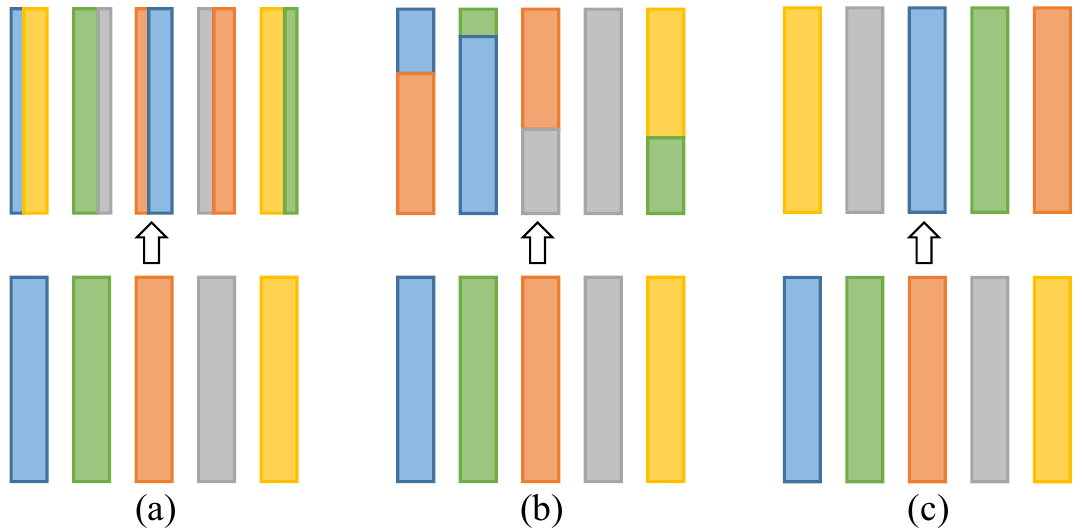}
   \caption{\textbf{Noise simulation strategies.} From left to right, the strategies include weighted averaging, random cropping coupled with concatenation, and random shuffling. Different instance queries are distinguished by different colors.
   }
   \label{fig:noise}
\end{figure}

\cref{fig:noise} illustrates the three noise simulation strategies proposed in this study, where the queries of the input instances $ \left\{ Q^i \in \mathcal{R}^{C}| i \in [1,N] \right\}$ are subjected to noise, resulting in the obtained queries of the output instances  $ \left\{ \hat{Q}^i \in \mathcal{R}^{C}| i \in [1,N] \right\}$. The weighted averaging strategy involves randomly combining each instance query with another randomly selected instance query:
\begin{equation}
  \left\{
    \begin{aligned}
	\mathcal{F}_{w}(Q) &=  \left\{ f_{w}({Q}^i)| i \in [1,N] \right\} \\
	f_{w}({Q}^i) &= \alpha * Q^{i} + (1 - \alpha) * Q^{j} \\
	\alpha = rand&(0, 1),\quad j = randint(0, N) \\
    \end{aligned}
    \right.
\end{equation}
The random cropping coupled with concatenation strategy is applied to perform both random cropping and concatenation on the instance query:
\begin{equation}
  \left\{
    \begin{aligned}
	\mathcal{F}_{c}(Q) &=  \left\{ f_{c}({Q}^i)| i \in [1,N] \right\} \\
	f_{c}({Q}^i) &= concatenate(Q^{i}[:k], Q^{j}[k:]) \\
	k = ra&ndint(0, C),\quad j = randint(0, N) \\
    \end{aligned}
    \right.
\end{equation}
The random shuffling strategy involves the random permutation of instance queries:
\begin{equation}
  \left\{
    \begin{aligned}
	\mathcal{F}_{s}(Q) &=  \left\{ {Q}^{\sigma(i)}| i \in [1,N] \right\} \\
	\sigma(i) &= shuffle([1, N])[i] \\
    \end{aligned}
    \right.
\end{equation}

\begin{figure}[t]
  \centering
   \includegraphics[width=1\linewidth]{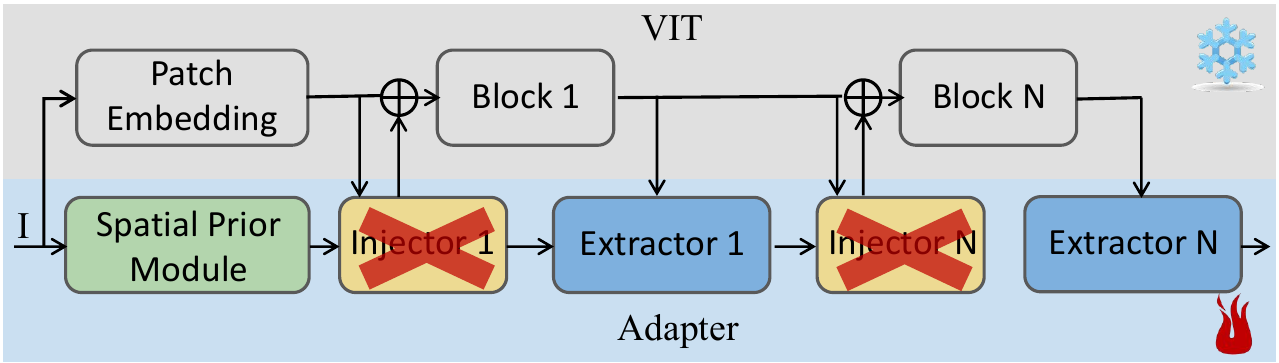}
   \caption{\textbf{Compact version of VIT-Adapter.} VIT has been entirely frozen, and all injectors in the Adapter have been removed.
   }
   \label{fig:vit_adapter}
\end{figure}
\begin{table*}[t]
\centering
\begin{tabular}{l|l|c|ccccc|ccccc}
	Phase & Method & mAP & mAP$^{S}$ & AP$_{50}^{S}$ & AP$_{75}^{S}$ & AR$_{1}^{S}$ & AR$_{10}^{S}$ & mAP$^L$ & AP$_{50}^{L}$ & AP$_{75}^{L}$ & AR$_{1}^{L}$ & AR$_{10}^{L}$  \\
\hline
	\multirow{5}{*}{Development}& Ours & \textbf{57.9} & \textbf{64.6} & \textbf{86.8} & \textbf{72.1} & \textbf{49.2} & \textbf{69.1} & \textbf{51.2} & 72.7 & \textbf{54.5} & 39.3 & 55.7 \\
	&jinyan & 54.0 & 58.9 & 78.6 & 66.1 & 45.1 & 63.8 & 49.0 & 69.0 & 54.4 & 38.8 & 54.2 \\
	&KainingYing & 53.8 & 61.2 & 83.5 & 68.8 & 48.0 & 65.7 & 46.4 & \textbf{74.2} & 41.5 & 39.4 & \textbf{55.8} \\
	&DeshuiMiao & 53.0 & 59.1 & 80.5 & 65.2 & 48.5 & 63.8 & 46.9 & 71.2 & 49.3 & \textbf{41.5} & 49.4 \\
	&SamsungMSL & 52.4 & 59.1 & 82.0 & 67.0 & 47.5 & 64.9 & 45.7 & 71.1 & 45.9 & 36.2 & 55.0 \\
\hline
	\multirow{5}{*}{Test} & Ours & \textbf{56.0} & \textbf{62.4} & \textbf{82.9} & \textbf{69.2} & \textbf{50.1} & \textbf{67.8} & \textbf{49.7} & \textbf{71.2} & \textbf{51.8} & 37.0 & \textbf{54.8} \\
	& KainingYing & 53.0 & 59.5 & 81.1 & 64.3 & 48.1 & 65.6 & 46.4 & 71.0 & 47.5 & 36.5 & 52.6 \\
	& GXU & 52.5 & 58.2 & 80.3 & 63.7 & 48.0 & 63.8 & 46.7 & 68.2 & 48.9 & 38.2 & 53.7 \\
	&guojuan & 52.3 & 58.3 & 80.0 & 63.7 & 48.2 & 64.0 & 46.3 & 67.0 & 48.9 & 38.1 & 53.1 \\
	&jmy & 52.1 & 58.3 & 79.9 & 63.8 & 48.2 & 64.2 & 45.9 & 67.5 & 49.0 & \textbf{38.4} & 52.5 \\
\hline
 \end{tabular}
\caption{\textbf{Leaderboards of the 5th LSVOS challenge.} ``mAP$^{S}$" represents the mean average precision (mAP) accuracy for short videos, while ``mAP$^{L}$" represents the mAP accuracy for long videos.}
 \label{tab:main_experiment}
\end{table*}

\begin{table}[t]
\centering
\begin{tabular}{l|c|cccc}
 Noise & Iter Number & AP & AP$_{l}$ & AP$_{m}$ & AP$_{h}$ \\
\hline
 None & 40k & 30.5 & 46.6 & 34.7 & 13.3 \\
 W. A. & 40k & 31.7 & 48.3 & 37.6 & 13.7 \\
 C. \& C. & 40k & 31.9 & \textbf{48.6} & 38.1 & 13.9 \\
 Shuffling & 40k & \textbf{32.7} & 48.5 & \textbf{38.9} & \textbf{14.4} \\
\hline
 None & 160k & 30.6 & 46.4 & 34.9 & 13.4 \\
 Shuffling & 160k & \textbf{34.3} & \textbf{50.4} & \textbf{41.0} & \textbf{15.8} \\
\hline  
 \end{tabular}
\caption{\textbf{Results of different noise simulation strategies on the OVIS validation dataset, using the original DVIS with R50 backbone as baseline.} ``None" indicates no noise added to the input, ``W. A." refers to weighted averaging, and ``C. \& C." denotes random cropping coupled with concatenation.}
 \label{tab:noise}
\end{table}

\begin{table}[t]
\setlength{\tabcolsep}{1.4mm}
\centering
\begin{tabular}{l|c|cccc}
 Backbone & COCO AP & AP & AP$_{l}$ & AP$_{m}$ & AP$_{h}$ \\
\hline
 Swin-L & 50.1 & 48.6 & 68.5 & 56.0 & 25.9 \\
 VIT-L(DINO v2) & 50.2 & 53.9 & 71.6 & 59.7 & 32.6 \\
\hline
 \end{tabular}
\caption{\textbf{The result of vision foundation model on OVIS validation dataset.}}
 \label{tab:vfm}
\end{table}
\subsection{Vision Foundation Model} \label{sec:vfm}

The visual foundation models have demonstrated impressive performance and generalization capabilities. Among them, the VIT-L model pretrained with DINO v2 has shown satisfying results in semantic segmentation and semantic matching without any fine-tuning. Video instance segmentation requires not only powerful segmentation abilities but also accurate instance matching capabilities, making the versatility showcased by DINO v2 particularly attractive. In this report, we introduce the DINO v2 pretrained visual foundation model to DVIS and investigate its impact on video instance segmentation.

Due to VIT's lack of capability to directly generate multi-scale features essential for dense prediction tasks, the VIT-Adapter \cite{vit-adapter} has been employed to address this limitation. However, the usage of VIT-Adapter comes at the cost of consuming a significant amount of GPU memory. In order to reduce the model's resource requirements, certain components were removed. As shown in \cref{fig:vit_adapter}, all injectors were eliminated, and the pre-trained VIT with DINO v2 was frozen, resulting in significant savings in GPU memory.

\section{Experiments}

Unless otherwise specified, we used the same training settings as DVIS. When using the DINO v2 pretrained VIT-L as the backbone, we first pre-trained the segmenter (Mask2Former \cite{Mask2Former}) on the COCO \cite{coco} dataset, using the same training settings as Swin-L \cite{swin}. For training DVIS on video datasets, the shortest edge of the input videos was randomly scaled to [288, 320, 352, 384, 416, 448, 480, 512], while during testing, the input videos were scaled to 360p.

\subsection{Main Result}
\cref{tab:main_experiment} illustrates the comparative results of different approaches, showcasing the superior performance of our method in both the development and test phases. Our approach demonstrates a significant advantage, outperforming the 2nd method by 3.9 AP and 3.0 AP in the development and test phases, respectively.

\subsection{Ablation Study}
\textbf{Denosing training strategy.}
\cref{tab:noise} presents the impact of the denoising training strategy on the OVIS dataset using the original DVIS \cite{DVIS} with a ResNet50 \cite{resnet} backbone as the baseline. The utilization of the random shuffling strategy resulted in an AP of 32.7, showcasing a performance gain of 2.2 AP. Additionally, by incorporating a higher number of training iterations, the random shuffling strategy achieved an AP of 34.3, surpassing the performance of the training strategy without denoising, which only attained an AP of 30.6.

\textbf{Vision Foundation Model.} 
The results of pretrained VIT-L using DINO v2 are presented in \cref{tab:vfm}. The visual foundation model does not improve the segmentation performance of the segmenter (50.2 vs. 50.1), but significantly enhances the performance of DVIS in video instance segmentation tasks (53.9 vs. 48.6). Experimental results demonstrate that the utilization of the visual foundation model leads to an increase of 3.1 AP for lightly occluded objects, 3.7 AP for moderately occluded objects, and 6.7 AP for heavily occluded objects. These results indicate that the introduction of the visual foundation model primarily enhances the instance tracking ability of DVIS rather than the segmentation ability.

\section{Conclusion}
In this report, we propose a denoising training strategy and introduce three noise simulation strategies that significantly improve the performance of DVIS. Additionally, we investigate the impact of incorporating a visual foundation model on the task of video instance segmentation. By combining an effective denoising training strategy and the visual foundation model, DVIS achieves the championship in the video instance segmentation track of the 5th LSVOS challenge at ICCV 2023, outperforming other methods by a large margin.

{\small
\bibliographystyle{ieee_fullname}
\bibliography{egbib}
}

\end{document}